\documentclass[11pt]{article}

\usepackage[preprint]{acl}

\usepackage{times}
\usepackage{latexsym}
\usepackage[T1]{fontenc}
\usepackage[utf8]{inputenc}
\usepackage{microtype}
\usepackage{inconsolata}
\usepackage{graphicx}
\usepackage{booktabs}
\usepackage{amsmath}
\usepackage{amssymb}
\usepackage{enumitem}
\usepackage{listings}
\usepackage{algorithm}
\usepackage{algpseudocode}
\usepackage{tikz}
\usetikzlibrary{arrows.meta,positioning,fit,backgrounds,calc,shapes.geometric}

\definecolor{codekw}{rgb}{0.13,0.29,0.53}
\definecolor{codecomment}{rgb}{0.40,0.45,0.40}
\definecolor{codestr}{rgb}{0.55,0.20,0.20}
\definecolor{codedeco}{rgb}{0.72,0.45,0.0}   
\definecolor{codebg}{gray}{0.96}             
\lstdefinestyle{cafe}{
  language=Python, basicstyle=\ttfamily\scriptsize,
  keywordstyle=\color{codekw}\bfseries, commentstyle=\color{codecomment}\itshape,
  stringstyle=\color{codestr}, showstringspaces=false, columns=fullflexible,
  breaklines=true, aboveskip=6pt, belowskip=3pt,
  backgroundcolor=\color{codebg}, xleftmargin=6pt, xrightmargin=4pt,
  framexleftmargin=6pt, framextopmargin=3pt, framexbottommargin=3pt,
  moredelim=[l][\color{codedeco}]{@}          
}

\newcommand\blfootnote[1]{%
  \begingroup
  \renewcommand\thefootnote{}\footnote{#1}%
  \addtocounter{footnote}{-1}%
  \endgroup
}

\newskip\sectionpostskip
\newskip\subsectionpostskip
\newskip\paragraphpreskip
\makeatletter
\def\section{\@startsection {section}{1}{\z@}{-1.0ex plus -0.3ex minus -.2ex}%
  {\sectionpostskip}{\large\bfseries\raggedright}}
\def\subsection{\@startsection{subsection}{2}{\z@}{-0.9ex plus -0.3ex minus -.2ex}%
  {\subsectionpostskip}{\normalsize\bfseries\raggedright}}
\def\paragraph{\@startsection{paragraph}{4}{\z@}%
  {\paragraphpreskip}{-1em}{\normalsize\bfseries}}
\makeatother
\AtBeginDocument{\raggedbottom}

\title{CAFE: A \textbf{C}ompound-\textbf{A}I \textbf{F}actorial \textbf{E}valuation Framework}

\author{%
  Fabian Lukassen\textsuperscript{1} \quad
  Christoph Weisser\textsuperscript{2} \quad
  Thomas Kneib\textsuperscript{1} \quad
  Alexander Silbersdorff\textsuperscript{1} \\
  \textsuperscript{1}University of Göttingen \quad
  \textsuperscript{2}Bielefeld University of Applied Sciences and Arts (HSBI) \\
  \texttt{fabian.lukassen@uni-goettingen.de} \quad
  \texttt{christoph.weisser@hsbi.de} \\
  \texttt{tkneib@uni-goettingen.de} \quad
  \texttt{asilbersdorff@uni-goettingen.de}
}

\begin{document}
\maketitle

\blfootnote{\noindent\hspace*{-2em}\begin{minipage}{\linewidth}\tiny
  Landing page: \url{https://cafe-ai.de} \\
  Demo: \url{https://cafe-ai.de/demo} \\
  Code: \url{https://github.com/fabian-lu/Cafe} \\
  Documentation: \url{https://fabian-lu.github.io/Cafe}
\end{minipage}}

\vspace{-2.8\baselineskip}
\begin{abstract}
We introduce \textbf{CAFE} (\textbf{C}ompound-\textbf{A}I \textbf{F}actorial \textbf{E}valuation), an
open-source platform that brings \emph{design of experiments} to the evaluation of compound AI
systems (CAIS). Such systems expose many interchangeable choices -- e.g. which retriever, model, or
prompt -- and practitioners rarely know which of them most affects answer quality. With CAFE, a
practitioner registers each swappable component of a pipeline as a factor to build
a factorial design over the chosen factors, run the resulting configurations, and
score the answers on a shared rubric using a configurable LLM judge together with human raters. From
these ratings it attributes answer-quality variance to the components and their interactions with
mixed-effects models and reports effect sizes, significance, the best configuration, cost
and latency trade-offs, and judge--human reliability. Whereas existing tools mostly either
\emph{search} for a good configuration or \emph{score} outputs in isolation, CAFE also
\emph{explains} which component drives quality and whether an observed difference is significant. We validate
CAFE on a retrieval-augmented question-answering (QA) pipeline over the HotpotQA benchmark dataset, where it recovers planted factor effects and stays calibrated under a permutation null.
CAFE is released as a
Python package and as a Web application.
\end{abstract}

\section{Introduction}
\label{sec:intro}

Modern AI applications rarely rest on a single language model. The current predominant pattern is the
\emph{compound AI system}~\citep{zaharia2024compound,chen2025standalone}: an application that
solves a task with several interacting components such as retrieval, reranking, context
assembly, prompting, one or more model calls, tool use, and guardrails. Each component can be realized by several competing techniques, retrieval alone might be sparse keyword
search, dense-embedding retrieval, or a reranking pipeline; any
other component can be swapped just as freely. Because these discrete, often non-differentiable choices combine, even a modest pipeline induces a
large, combinatorial space of candidate \emph{configurations}. The growing literature helps practitioners
\emph{build} and \emph{optimize} such pipelines, from declarative programming
models~\citep{khattab2024dspy} to end-to-end configuration
search~\citep{kartal2025ragsmith,lee2025cais}.

However, choosing which configuration to ship raises questions that optimizers only answer indirectly. 
Suppose that the builder of a retrieval-augmented assistant can pick
among three base models, two retrieval corpora, and three prompts: 18 candidate configurations.
The practitioner's immediate goal is the single best-performing one, which CAFE
reports directly; but a trustworthy, actionable decision rests on two coupled questions. First,
\emph{which component drives answer quality and where is effort best spent?} A team's time is
finite, and it matters whether the next gain will come from engineering the prompt or from
building a stronger retrieval pipeline. Components also interact, often non-monotonically -- a
stronger model may help only when retrieval is good, and adding model calls can help some queries
while hurting others~\citep{chen2024morecalls} -- so a disjoint perspective risks to mismeasure
the system by neglecting such interactions. Second,
\emph{is an observed difference significant?} A gap between two configurations can be an artifact of
chance rather than a genuine effect, and comparing averaged scores from a single run of each
gives no way to tell the two apart. An optimizer returns the best-scoring configuration, but not a calibrated account of
\emph{why} it wins or \emph{whether} the margin is trustworthy.

To address these issues CAFE employs \emph{design of experiments} (DoE) and mixed-effects
models \citep{fisher1935design,montgomery2017doe,pinheiro2000mixed}.
A factorial design estimates each component's effect \emph{and} the interactions among
components from a balanced set of runs; replicating runs and modeling the outcome
with random effects separates real differences both from noise and from the fact that some
inputs are simply harder than others. However, there is currently no usable system that packages this methodology
for people who build compound AI pipelines.

CAFE is an open-source evaluation framework
with four contributions: (i) a \textbf{pipeline-as-factors abstraction} that surfaces any
swappable component as a \emph{factor} whose alternative implementations become its
\emph{levels}; (ii) a \textbf{design-and-execution engine} that builds a factorial design over the chosen factors,
runs each configuration against the target system; (iii) a \textbf{dual-rating workflow} in which the same answers are scored
on a shared rubric by a fully configurable LLM judge and by humans, with inter-rater
reliability reported between them; and (iv) a \textbf{variance-attribution layer} that fits
mixed-effects models matched to the rating scale, reports how much each component and interaction
contributes with effect sizes and significance, and identifies the best configuration.

CAFE is released as an open-source early-stage 
\href{https://github.com/fabian-lu/Cafe}{Python package} (licensed under Apache~2.0) with a companion web application (a screenshot of the web interface is in
Appendix~\ref{app:supp}); we also provide
\href{https://fabian-lu.github.io/Cafe/notebooks/01_quickstart/}{example notebooks} for system demonstration. The project welcomes any
\href{https://github.com/fabian-lu/Cafe/blob/main/CONTRIBUTING.md}{contributions}, where 
\href{https://cafe-ai.de/roadmap.html}{roadmap} lists planned features, a
selection of which we also highlight in Section~\ref{sec:limitations}.

\begin{table}[H]
\centering
\setlength{\tabcolsep}{4pt}
\renewcommand{\arraystretch}{1.05}
\resizebox{\columnwidth}{!}{%
\begin{tabular}{lccccc}
\toprule
System & Pipeline-level & DoE / Variance attrib. & Human + judge & Replication & Config.\ search \\
\midrule
RAGAS        & part        & --          & --          & --          & --          \\
ARES         & part        & --          & --          & --          & --          \\
DeepEval     & \checkmark  & --          & part        & --          & --          \\
TruLens      & \checkmark  & --          & part        & --          & --          \\
LangSmith    & \checkmark  & --          & \checkmark  & --          & --          \\
Inspect AI   & \checkmark  & --          & part        & part        & --          \\
DSPy         & \checkmark  & --          & --          & --          & \checkmark  \\
RAGSmith     & \checkmark  & --          & --          & --          & \checkmark  \\
\textbf{CAFE}& \checkmark  & \checkmark  & \checkmark  & \checkmark  & --          \\
\bottomrule
\end{tabular}%
}
\setlength{\abovecaptionskip}{2pt}
\caption{\footnotesize CAFE versus representative systems.}
\label{tab:compare}
\end{table}

\section{Related Work}
\label{sec:related}

The evaluation of CAIS touches on three lines of work: tools that \emph{optimize} pipelines,
tools that \emph{score} their outputs, and the statistics of AI evaluation. Table~\ref{tab:compare}
contrasts CAFE with representative evaluation systems.

\paragraph{Optimizing compound systems.}
One line of work \emph{searches} the configuration space for a high-scoring pipeline: DSPy compiles
and tunes declarative pipelines~\citep{khattab2024dspy}, TextGrad backpropagates textual
feedback~\citep{yuksekgonul2025textgrad}, LLMSelector allocates models to
modules~\citep{chen2025llmselector}, Optimas aligns local rewards with global
performance~\citep{wu2025optimas}, and RAGSmith and multi-objective search explore large RAG
spaces~\citep{kartal2025ragsmith,barker2025faster}.

\paragraph{Evaluation frameworks and tools.}
Holistic benchmarks such as HELM~\citep{liang2023helm} characterise \emph{models} across scenarios
rather than \emph{pipelines} across component choices. Metric libraries score pipeline outputs with
LLM judges: RAGAS~\citep{es2024ragas} and ARES~\citep{saadfalcon2024ares}, which adds
prediction-powered confidence
intervals~\citep{fisch2024stratified}, alongside DeepEval~\citep{deepeval}, a \texttt{pytest}-style
framework offering G-Eval~\citep{liu2023geval} and component-level metrics, and TruLens~\citep{trulens}, which computes
feedback functions over execution traces. Execution and annotation platforms (LangSmith~\citep{langsmith},
Inspect~AI~\citep{inspectai}, and Argilla~\citep{argilla}) supply run harnesses, judge scoring, and
human annotation queues.

\paragraph{Statistics for AI evaluation.}
A growing literature views evaluation through the lense of statistical inference: quantifying benchmark
uncertainty~\citep{wang2025reprorag}, hierarchical Bayesian modeling~\citep{luettgau2025hibayes},
measurement error in LLM pipelines~\citep{messing2026hidden}, and ordinal rather than interval models
for rubric ratings~\citep{howcroft2021ordinal,taylor2023clmm,syiem2026ordinal}; variance-based
sensitivity analysis is the classical instrument for apportioning output variance to
inputs~\citep{saltelli2008primer}. The two closest works are \citet{haase2026variability}, who
partition output variance into prompt, model and sampling components but for standalone creative
generation with a linear model and a single automatic rater, and \citet{mustahsan2025icc}, who
summarise run-to-run inconsistency with a single intraclass correlation; neither attributes variance
across a designed factorial spanning the full pipeline. Recent work also attributes faults or
perturbations within compound systems~\citep{chowdhury2026seta,nilayam2026quiver}, while a parallel
effort calibrates and audits LLM
judges~\citep{gu2024judgesurvey,han2025verdict,boyeau2025autoeval,park2025rautoeval,dubois2025skewed,guerdan2025indeterminacy}
and documents the nondeterminism that motivates
replication~\citep{atil2024nondeterminism,haldar2025roulette,bjarnason2026randomness}. Factorial
studies of LLM outputs are beginning to appear~\citep{vazquez2025factorial,lukassen2026xai}.

\paragraph{The gap.}
No existing system combines these threads (Table~\ref{tab:compare}). Optimizers return the best-scoring
configuration but not a calibrated account of \emph{why} it is best. Metric libraries and execution platforms score outputs and gather
ratings, but leave the experimental design and statistical analysis to the user. The statistical
methods above are directly relevant, yet are not packaged as a usable tool. Section~\ref{sec:framework} now describes how CAFE operationalizes this gap as a framework.

\section{The CAFE Framework}
\label{sec:framework}

{%
\setlength{\abovecaptionskip}{1pt}%
\setlength{\belowcaptionskip}{0pt}%
\setlength{\intextsep}{2pt}%
\begin{figure}[H]
  \centering
  \includegraphics[width=\columnwidth]{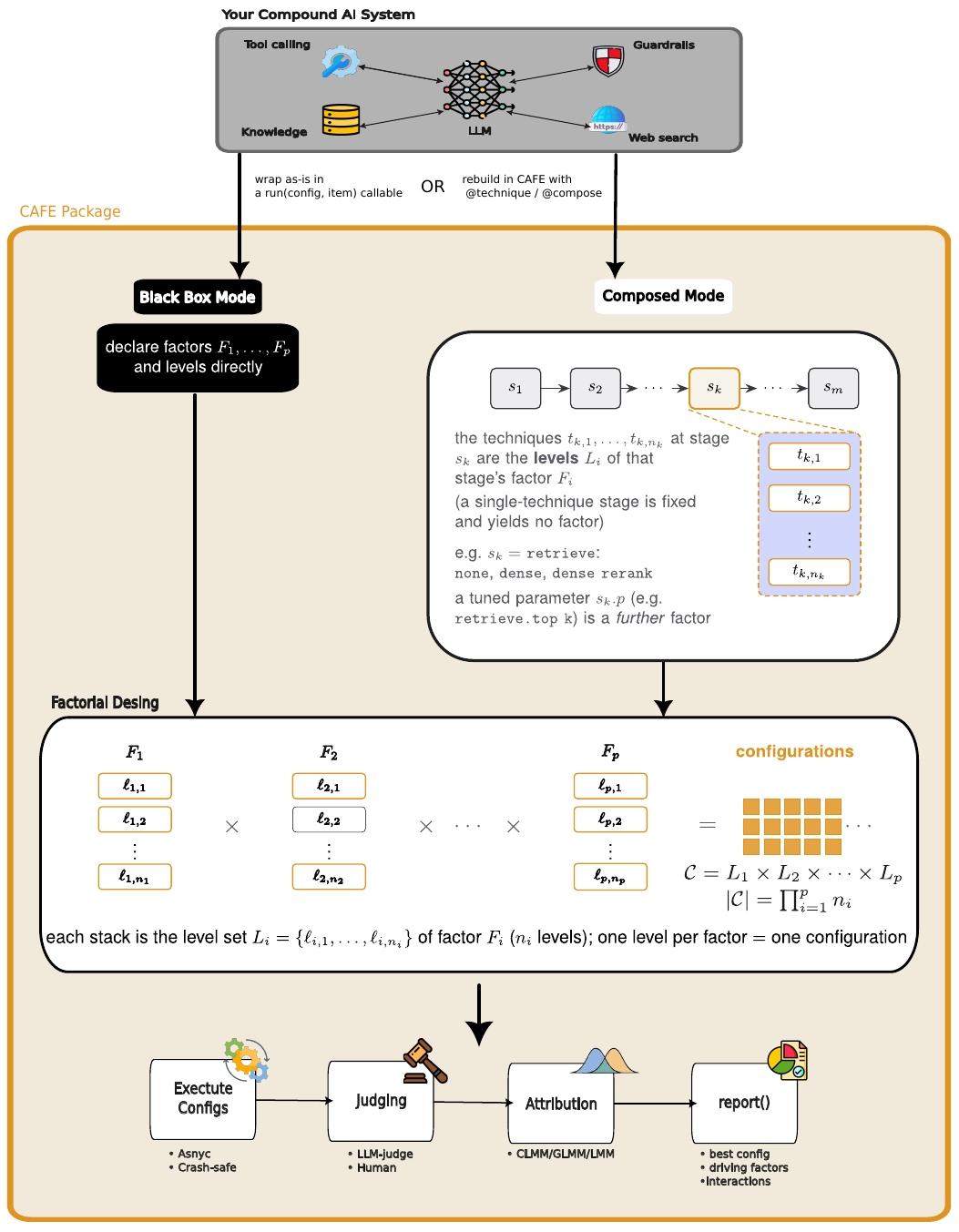}
  \caption{\footnotesize CAFE architecture.}
  \label{fig:arch}
\end{figure}%
}

CAFE turns the choices in a CAIS into experimental \emph{factors} (Figure~\ref{fig:arch}). A study varies factors
$F_1,\dots,F_p$, each ranging over a finite set of \emph{levels} $L_i$; choosing one level per factor
gives a \emph{configuration} $c=(c_1,\dots,c_p)\in\mathcal{C}=L_1\times\cdots\times L_p$, and the
\emph{full factorial} is this entire product. Running $c$ on an input $x\in\mathcal{X}$ yields an output
$\mathrm{Run}(c,x)$ that a rubric maps to a quality rating $Y$; the system and judge being
nondeterministic, $Y$ is a random variable, since the same $c$ can score differently across runs. Two
kinds of choice become factors: \emph{which component} performs a step -- e.g.\ the retriever, with
levels \texttt{none}, \texttt{dense}, \texttt{dense\_rerank} -- and a \emph{parameter} of a component,
such as the decoding temperature.

When a pipeline is built inside CAFE, these factors are read off its structure: the system is an
ordered sequence of \emph{stages} $s_1,\dots,s_m$ (each a point where a choice is made), and the
\emph{techniques} registered for a stage $s_k$ are the levels of that stage's factor, while a technique's tunable parameter $s_{k.p}$
(e.g.\ \texttt{retrieve.top\_k}) is a further factor.

Given factors and inputs, CAFE casts the questions of Section~\ref{sec:intro} as estimation problems
on $Y$: \emph{attribution} -- how much of the variation in $Y$ each factor $F_i$ and each interaction
explains; \emph{selection} -- the configuration $\arg\max_{c}\mathbb{E}[Y\mid c]$; and
\emph{significance} for both, with valid uncertainty under the run-to-run variation of $\mathrm{Run}$.
Algorithm~\ref{alg:study} formalises this procedure. We describe each part
in turn.

{\setlength{\intextsep}{2pt}%
\begin{algorithm}[H]
\small
\caption{A CAFE study}
\label{alg:study}
\begin{algorithmic}[1]
\State declare factors $F_1,\dots,F_p$ and their levels $L_1,\dots,L_p$
\State $\mathcal{C} \gets L_1 \times \dots \times L_p$ \Comment{full factorial: every configuration}
\ForAll{configuration $c \in \mathcal{C}$, input $x \in \mathcal{X}$, replicate $r$}
  \State $a_{cxr} \gets \Call{Run}{c, x}$ \Comment{execute; concurrency-bounded, resumable}
  \State $y_{cxr} \gets \Call{Rate}{a_{cxr}}$ \Comment{LLM judge and/or human, shared rubric}
\EndFor
\State fit the model matched to the rubric scale (CLMM / logistic / linear)
\State attribute variance; report effect sizes, significance, the best configuration, the cost frontier, and~$\alpha$
\end{algorithmic}
\end{algorithm}}

\subsection{Defining a system}
CAFE supports two ways to supply $\mathrm{Run}$ from Algorithm~\ref{alg:study}. In the common case, the pipeline is \emph{composed} in
CAFE itself: each technique is registered on its stage with a one-line decorator, and an orchestration
function wires the stages together (Listing~\ref{lst:pipeline}). CAFE then varies each factor over its
levels and, because it executes the pipeline, records per-stage latency and dollar cost with no
extra wiring. One level of a factor may be \texttt{None}, which tells CAFE to skip that stage entirely,
so a component can be switched on and off (e.g. running with or without a reranker). Alternatively, a system that cannot or need not be instrumented -- e.g.\ an external HTTP
endpoint -- is driven as a \emph{black box}: any callable \texttt{run(config, item)} is a system --
CAFE passes it a configuration and an input, and the callable reads each factor's chosen level from
\texttt{config} -- and its factors are declared directly. Black-box mode gives up only the per-stage
latency and cost breakdown, leaving the design, rating, and variance attribution unchanged.

\begin{lstlisting}[style=cafe,belowskip=0pt]
import cafe
pipe = cafe.Pipeline()

# `retrieve`: two techniques = the levels of the retrieve factor
\end{lstlisting}
\begin{lstlisting}[style=cafe,aboveskip=0pt,captionpos=b,caption={\footnotesize Composed mode: techniques are registered on stages with decorators and \texttt{@compose} wires
them into a pipeline; \texttt{cost\_usd} attaches a per-call price CAFE tracks. The same \texttt{Study}
runs unchanged in black-box mode over any \texttt{run(config, item)} callable.},label={lst:pipeline}]
@pipe.technique(stage="retrieve", name="none")
async def retrieve_none(ctx, query):
    return []                          # parametric memory only

@pipe.technique(stage="retrieve", name="dense", cost_usd=0.0004)
async def retrieve_dense(ctx, query, top_k=8):
    return dense_search(query, k=top_k)      # top_k -> a retrieve.top_k factor

# `generate`: the `model` keyword becomes the generate.model factor
@pipe.technique(stage="generate", name="llm")
async def generate(ctx, query, docs, model="gpt-oss-20b"):
    return await cafe.complete(model, prompt(query, docs))

@pipe.compose
async def rag(config, item, ctx):            # wire the stages together
    docs = await ctx.run("retrieve", query=item["text"])
    return await ctx.run("generate", query=item["text"], docs=docs)

study = cafe.Study(
    name="rag-study",
    system=pipe,
    factors=[pipe.factor("retrieve"),
             cafe.Factor("generate.model", ["gpt-oss-20b", "gpt-oss-120b"])],
    dataset=data, rubric=rubric, judge=judge, replications=2,
)
\end{lstlisting}

\subsection{Design, cost, and execution}
From the chosen factors, CAFE enumerates a \emph{full factorial} -- every combination of levels -- and
\texttt{cafe.preflight} reports the configuration count, an estimated compute time and dollar cost, and
a design-adequacy check (for instance, too few inputs for a stable random effect) before anything runs.
Execution then runs each pair $(c,x)\in\mathcal{C}\times\mathcal{X}$ under a configurable concurrency
bound and checkpoints incrementally, so a study interrupted after hours resumes rather than restarts.
Every cell is run with \emph{replication}: a compound system can return different outputs on repeated
calls even at temperature zero~\citep{atil2024nondeterminism}, and the repeats let the analysis
estimate within-cell variation instead of mistaking it for an effect. Per-stage latency and cost are
recorded throughout, feeding the multi-objective analysis of Section~\ref{sec:methods}.

\subsection{Rating}
Each answer is scored on a \emph{rubric} that fixes an \emph{ordinal}, \emph{binary}, or
\emph{numeric} scale. Two rater types apply the same rubric. An \emph{LLM judge} is fully
configurable -- its model, its prompt, and the rubric it applies are all chosen by the user, and it scores with or without a reference answer; CAFE ships
prompt presets and stores the exact prompt and
raw response behind every verdict for auditing. Repeated judge passes are retained so that judge
self-inconsistency can be measured~\citep{haldar2025roulette}. \emph{Humans} rate through a
fillable answer sheet (\texttt{cafe.answer\_sheet} exports it as csv,
\texttt{cafe.human\_ratings} reads it back), thus allowing reliability analysis between them.

\subsection{Statistical analysis}
\label{sec:methods}
The statistical analysis is CAFE's analytical core, reported in three layers, and it is deliberately driven by the rubric:
the declared scale type selects the model, so each response is analysed on the scale it lives on rather
than coerced to an interval mean.

\emph{Descriptive.} Per-level marginal means, per-configuration means, and the best configuration give
the immediate picture of the design.

\emph{Scale-matched inference.} Write $Y$ for an answer's rating and $y_{cxr}$ for its value under
configuration $c$ on input $x$, replicate $r$, with $c$ summarised by its factor-level design vector
$\mathbf{z}_c$ (main effects and, by default, two-way interactions), and let
$\eta = \mathbf{z}_c^{\top}\boldsymbol{\beta} + u_x$ be the linear predictor, where
$u_x \sim \mathcal{N}(0,\sigma^2)$ is a per-input random intercept. The declared scale selects one of
three mixed models:
{\setlength{\abovedisplayskip}{3pt}\setlength{\belowdisplayskip}{3pt}%
\setlength{\abovedisplayshortskip}{2pt}\setlength{\belowdisplayshortskip}{2pt}%
\setlength{\jot}{1pt}%
\begin{equation*}
\begin{aligned}
\text{ordinal (CLMM):}\quad & \operatorname{logit}\Pr(Y \le q) = \theta_q - \eta,\\
\text{binary (GLMM):}\quad  & \operatorname{logit}\Pr(Y = 1) = \eta,\\
\text{numeric (LMM):}\quad  & Y = \eta + \varepsilon.
\end{aligned}
\end{equation*}}
The ordinal case is a \emph{cumulative link mixed
model}~\citep{mccullagh1980,agresti2002,taylor2023clmm}; unlike the binary and numeric models it has
no fixed intercept, as its $K-1$ ordered thresholds $\theta_1 < \dots < \theta_{K-1}$ play that role
(which keeps $\mathbf{z}_c^{\top}\boldsymbol{\beta}$ identifiable). All three share the per-input \emph{random} intercept
$u_x$, which absorbs the fact that some inputs are simply harder, so factor effects are estimated
\emph{within} input. Because Python has no
standard fitter for the ordinal and binary models, CAFE fits them by calling \textsf{R} in a
subprocess, \texttt{ordinal::clmm}~\citep{christensen2023ordinal} and
\texttt{lme4::glmer}~\citep{bates2015lme4}, passing the data as a table and reading the 
estimates. The linear model is
fit natively.

\emph{Effect sizes and attribution.} Alongside the scale-matched model, a linear mixed model gives an
\emph{approximate}, scale-agnostic effect-size view: a Type-II analysis of
variance~\citep{langsrud2003anova} gives each factor an $F$-test and a partial $\eta^2$~\citep{cohen1988},
a model $R^2$~\citep{nakagawa2013r2}, and pairwise contrasts give Cohen's $d$~\citep{cohen1988}. Ranking
factors by partial $\eta^2$ is a variance attribution~\citep{saltelli2008primer} that answers ``which
component drives quality,'' and the best end-to-end configuration is read from the fitted means. Because
it treats the scale as interval, this view is a standardized summary for ranking and cross-study
comparison rather than the primary inference (Section~\ref{sec:limitations}).

\emph{Reliability and robustness.} Because the quality signal is itself produced by raters, CAFE
computes Krippendorff's~$\alpha$~\citep{krippendorff2004content} between the LLM judge and the humans, and among the humans, with the distance metric matched to the scale (ordinal, nominal, or
interval). Finally, because
latency and cost are recorded, CAFE reports the \emph{Pareto frontier}~\citep{barker2025faster} over quality, cost, and latency,
exposing a configuration that is marginally better, but far slower or more expensive.

\subsection{Implementation and availability}
CAFE is distributed as a Python package with modules for design, execution, judging, and statistics.
A web application exposes the same engine over a FastAPI (JSON) API that a React front-end consumes,
backed by a PostgreSQL store; studies run as in-process asynchronous task, and the
whole stack is deployed with Docker Compose. The app discovers the available systems from a configured
techniques directory, where \texttt{@compose} and \texttt{@technique} registrations populate its factor/level menus.

\section{Case Study and Evaluation}
\label{sec:eval}

We validate CAFE against \textbf{known answers}: we run it on a real compound system built so that
certain effects hold \emph{by construction}, then check whether CAFE recovers them (the full study is
a runnable \href{https://fabian-lu.github.io/Cafe/notebooks/evaluation/evaluation/}{notebook}).
Concretely, CAFE must \emph{recover} an effect we know is present while ignoring one we know is absent
(E1, construct validity); \emph{not invent} significance once we destroy all real structure
(E2, calibration); \emph{agree with human} raters, since the whole analysis rests on the judge's scores
(E3); and select a best configuration that \emph{generalizes} beyond the exact questions asked (E4).

\paragraph{The task.}
The target system is a retrieval-augmented QA pipeline evaluated on
\textbf{HotpotQA}~\citep{yang2018hotpotqa}, a multi-hop
question-answering benchmark in which answering a question requires combining facts from two different
Wikipedia \emph{paragraphs}. Each question comes with (i)~a short gold answer and (ii)~ten candidate paragraphs, of which
two are the \emph{gold} paragraphs that contain the supporting facts and eight are \emph{distractors}.
 Thus, we know both the correct answer and the relevant evidence. We keep the hard, multi-hop,
non-yes / no questions (yes / no answers are guessable and would blunt the retrieval signal) and sample
$n = 50$.

\paragraph{Screening.}
Because a question's own ten paragraphs make retrieval trivial, we pool
\emph{every} question's paragraphs into one shared $\sim$500-document corpus, embedded with
\texttt{bge-m3}: each question's two gold paragraphs must now be found among $\sim$500. Before spending
anything on the full study we \emph{screen} this corpus by measuring \emph{gold-paragraph recall@$k$}
-- how often the retriever's top-$k$ actually contains a question's gold paragraphs. Recall is $0.76$
at $k{=}2$, $0.87$ at $k{=}4$, and $0.97$ at $k{=}12$.
It confirms the gold evidence is retrievable, so a no-retrieval baseline is \emph{guaranteed} to be
worse, the known-signed anchor. Furthermore, it shows that plain top-2 retrieval misses the gold about a
quarter of the time, while the top-12 almost always contains it: that gap is precisely the room a
\emph{reranker} would need to help.

\paragraph{The factors.}
The system exposes three factors ($3\times3\times2 = 18$ configurations):
\begin{itemize}[wide=0pt, labelindent=0pt, itemsep=0pt, topsep=0pt, parsep=0pt]
\item \texttt{retrieve} $\in\{$none, dense, dense\_rerank$\}$ -- the \textbf{anchor}. \texttt{none}
  passes no evidence (the model answers from parametric memory); \texttt{dense} feeds the
  \texttt{bge-m3} top-2; \texttt{dense\_rerank} has an LLM reorder the top-12 down to 2, a
  RankGPT-style reranker~\citep{sun2023rankgpt}.
\item \texttt{finalize} $\in\{$on, off$\}$ -- the \textbf{negative control}. Both levels return the
  answer unchanged, so the true effect is exactly zero and CAFE must call it non-significant.
\item \texttt{generate.model} $\in\{$gemma3-4b, gpt-oss-20b, gpt-oss-120b$\}$ -- an \textbf{expected}
  effect: larger models should score higher~\citep{kaplan2020scaling,roberts2020knowledge}; open is \emph{by how much}, and whether strong retrieval
  makes model size matter less.
\end{itemize}

\paragraph{Scoring.}
Each of the $18 \times 50 = 900$ answers is graded on a three-level ordinal correctness rubric --
$0$ incorrect/off-topic, $1$ partially correct, $2$ fully correct (exact wording irrelevant) -- by a
\emph{separate}, strong judge model (DeepSeek-V4-Pro). The judge is shown the question, the \emph{gold reference answer}, and the
answer to grade, and is instructed to score factual correctness against the reference. The complete \texttt{report()} output of this study is in
Appendix~\ref{app:supp} (Listing~\ref{lst:report}).

\paragraph{E1: recovery of known effects.}
Reading Listing~\ref{lst:report} top to bottom, CAFE reproduces what we built in. The \textbf{anchor}
is recovered with the right order and dominance: the marginal mean quality rises \texttt{none}
$0.75 <$ \texttt{dense} $1.48 <$ \texttt{dense\_rerank} $1.71$, and \texttt{retrieve} is by far the
largest effect (partial $\eta^2 = 0.228$, $p < 10^{-3}$). The reranker's edge over plain \texttt{dense}
is small but significant ($d = 0.31$), as the recall screen suggested it would be. The \textbf{negative
control} \texttt{finalize} is null on every layer ($\eta^2 \approx 0$, $p = 0.68$), and
\texttt{generate.model} behaves as expected -- larger models score higher ($\eta^2 = 0.052$,
$p < 10^{-3}$). The best configuration is \texttt{dense\_rerank} $+$ \texttt{gpt-oss-120b} (mean
$1.78$), with \texttt{finalize} immaterial. Full statistics ($F$, Cohen's $d$, and CLMM log-odds) are
in Listing~\ref{lst:report}.

\paragraph{Interactions.}
The factorial also flags a significant \texttt{retrieve} $\times$ \texttt{generate.model} interaction
($p = 0.002$). The cell means show its shape: the model gap is widest with no retrieval
(\texttt{none}: gemma3-4b $0.36$ vs.\ gpt-oss-120b $1.13$) and narrows as retrieval improves
(\texttt{dense\_rerank}: $1.59$ vs.\ $1.78$). Equivalently, retrieval helps the \emph{weaker} model
most -- gemma3-4b gains $+1.23$ from \texttt{none} to \texttt{dense\_rerank}, gpt-oss-120b only $+0.65$
-- because with the evidence in context even a small model can read the answer off, whereas with none
it falls back on weaker parametric knowledge (the CLMM confirms the sign: \texttt{none}
$\times$ gpt-oss-120b $=+1.85$, $p < 10^{-3}$). A one-factor-at-a-time sweep cannot see this;
surfacing it is exactly what a factorial design is for.

\paragraph{E2: does CAFE invent effects?}
E2 is the converse of E1: does CAFE avoid \emph{inventing} effects that are not there? We build data in which, by construction, \emph{no} factor has any effect, randomly 
reshuffling the $900$ verdicts onto different answers so that every link between a factor and quality
is broken. A calibrated test should then flag factors at the nominal Type~I error rate, $\alpha = 0.05$; over $500$ independent
reshuffles CAFE flags them $4.9\%$ of the time -- indicating no spurious
significance.

\paragraph{E3: does the judge track human judgement?}
The whole analysis rests on the LLM judge's scores, so the final check is whether they agree with
people. Two human experts independently re-scored a stratified sample of $61$ answers, applying the same rubric \emph{blind} to the judge's verdicts. Measured with Krippendorff's ordinal
$\alpha$, the two raters agree strongly with \emph{each other} ($\alpha = 0.91$), and the LLM-judge agrees
with them about as well -- $\alpha = 0.95$ and
$0.89$, yielding $\alpha = 0.92$ across all three. Every value lands in the
conventional ``reliable'' band ($\alpha \ge 0.80$).

\paragraph{E4: does the winner survive new questions?}
Finally, is the best configuration a real winner or an artifact of these particular $50$ questions? Using cross-validation, we
repeatedly split the questions into two random halves, recompute the configuration ranking on each half
independently, and measure how well the two rankings agree. Across splits the results correlate strongly
(Spearman $\rho = 0.91$). Disagreement is only found among a cluster of near-tied top
configurations, the very gaps CAFE had already reported as non-significant.

\section{Conclusion}
\label{sec:conclusion}

CAFE brings design of experiments to the evaluation of compound AI systems, turning pipeline
components into factors and attributing answer-quality variance to them and their interactions with a
mixed-effects model matched to the rubric's scale. On a retrieval-augmented HotpotQA study it recovers
planted effects, calls a negative control null, stays calibrated under a permutation null, and agrees
with expert human raters. CAFE is open-source, and aims to make designed, statistically grounded
evaluation routine for the teams who build compound AI systems.

\section{Limitations and Future Work}
\label{sec:limitations}

CAFE's own limitations are statistical and computational. Its scale-agnostic effect-size view (partial
$\eta^2$, Cohen's $d$) treats the ordinal scale as an interval and is only an approximate ranking aid;
mixed-effects fits can be unstable on small or sparse designs, as the near-singular random effect in
our study shows; and full-factorial designs grow multiplicatively in cost. Planned extensions address
these: an order-independent \emph{deviance partition} that apportions explained deviance across
factors~\citep{gromping2006relaimpo}, fractional-factorial screening to cover large spaces at a
fraction of the runs, and Bayesian mixed models~\citep{burkner2017brms} for full posterior effect
sizes and credible intervals. Modelling \emph{institutional} rater effects -- how different peer groups
score the same system -- is a further direction~\citep{luettgau2025hibayes}.

One caveat concerns our \emph{evaluation} rather than CAFE itself: it validates the instrument on a
relatively easy task -- a three-level ($0$--$2$) correctness judgment against a gold answer is not
where LLM judges struggle, and a subjective rubric (tone, style, helpfulness) would likely show lower
judge--human agreement. Judging is where CAFE will also grow: multi-criteria scoring (several named
scores per answer) and, where a few human labels are available, prediction-powered inference to
\emph{debias} the judge's estimates against them~\citep{boyeau2025autoeval,fisch2024stratified,park2025rautoeval}.

\bibliography{references}

\clearpage
\onecolumn
\appendix

\section{Supplemental Material}
\label{app:supp}

\begin{figure}[H]
  \centering
  \caption{\footnotesize The CAFE web application (Results view) on the HotpotQA recovery study of
  Section~\ref{sec:eval}: the factors \texttt{retrieve} $\times$ \texttt{generate.model} $\times$ the
  \texttt{finalize} placebo (18 configurations, 50 questions, a 0--2 correctness rubric). It reports
  overall quality, the best configuration, per-factor attribution ($F$, $p$, partial $\eta^2$, and the
  share of variance explained), and per-level marginal means, all computed by \texttt{cafe-core} and
  served over the API. Listing~\ref{lst:report} is the same analysis as console output.}
  \label{fig:webui}
  \includegraphics[width=0.85\textwidth]{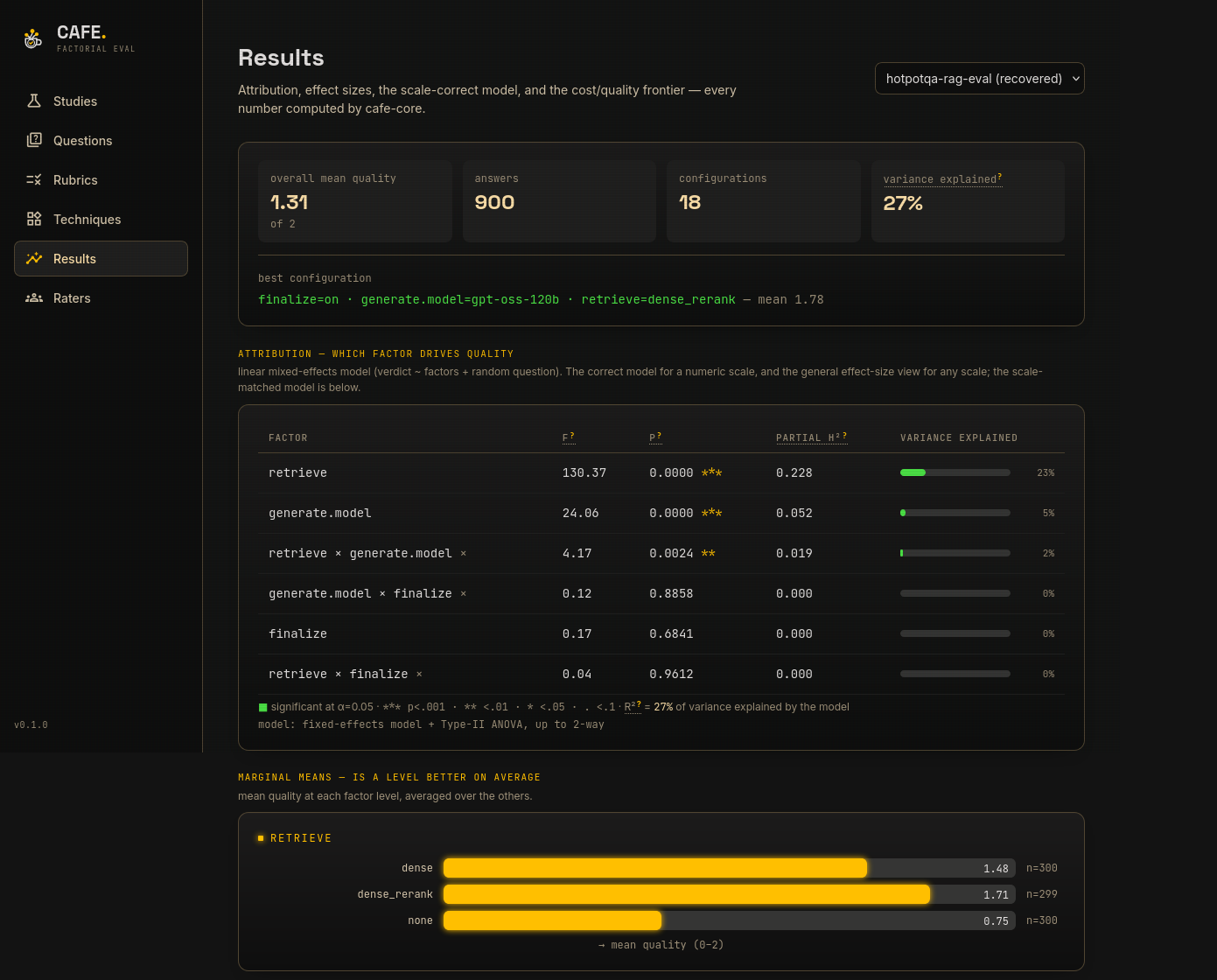}
\end{figure}

\begin{lstlisting}[basicstyle=\ttfamily\footnotesize,backgroundcolor=\color{codebg},columns=fixed,breaklines=true,xleftmargin=4pt,caption={\footnotesize Complete console output of \texttt{result.report()} on the study of Figure~\ref{fig:webui}: descriptive means (overall, ranked per configuration, and per factor) with the best configuration; the inferential linear mixed model (Type-II $F$, partial $\eta^2$, and pairwise Cohen's $d$) with its near-singular random-effect note; and the ordinal cumulative-link mixed model. Retrieval dominates and is correctly ordered (none $<$ dense $<$ dense\_rerank), the base model helps, and the placebo \texttt{finalize} is null: the recovery result of Section~\ref{sec:eval}.},label={lst:report}]
900 answers | 18 configs | 50 inputs | 900 ratings
best: finalize=on | generate.model=gpt-oss-120b | retrieve=dense_rerank
pipeline: 900 answers -> 900 judged (1 unparseable) -> 899 usable verdicts

============================================================
DESCRIPTIVE -- means & best configuration
============================================================
verdicts: 900 (899 usable)   factors: retrieve, generate.model, finalize
overall mean quality: 1.31  (n=899)

per-configuration mean quality:
  1.78  (n=50)  finalize=on  | generate.model=gpt-oss-120b | retrieve=dense_rerank
  1.78  (n=49)  finalize=off | generate.model=gpt-oss-120b | retrieve=dense_rerank
  1.76  (n=50)  finalize=on  | generate.model=gpt-oss-20b  | retrieve=dense_rerank
  1.74  (n=50)  finalize=off | generate.model=gpt-oss-20b  | retrieve=dense_rerank
  1.62  (n=50)  finalize=off | generate.model=gpt-oss-120b | retrieve=dense
  1.60  (n=50)  finalize=on  | generate.model=gpt-oss-120b | retrieve=dense
  1.60  (n=50)  finalize=on  | generate.model=gemma3-4b    | retrieve=dense_rerank
  1.58  (n=50)  finalize=on  | generate.model=gpt-oss-20b  | retrieve=dense
  1.58  (n=50)  finalize=off | generate.model=gemma3-4b    | retrieve=dense_rerank
  1.50  (n=50)  finalize=off | generate.model=gpt-oss-20b  | retrieve=dense
  1.32  (n=50)  finalize=off | generate.model=gemma3-4b    | retrieve=dense
  1.28  (n=50)  finalize=on  | generate.model=gemma3-4b    | retrieve=dense
  1.16  (n=50)  finalize=on  | generate.model=gpt-oss-120b | retrieve=none
  1.10  (n=50)  finalize=off | generate.model=gpt-oss-120b | retrieve=none
  0.80  (n=50)  finalize=on  | generate.model=gpt-oss-20b  | retrieve=none
  0.74  (n=50)  finalize=off | generate.model=gpt-oss-20b  | retrieve=none
  0.36  (n=50)  finalize=off | generate.model=gemma3-4b    | retrieve=none
  0.36  (n=50)  finalize=on  | generate.model=gemma3-4b    | retrieve=none

per-factor marginal means:
  retrieve:        none 0.75 (n=300)   dense 1.48 (n=300)   dense_rerank 1.71 (n=299)
  generate.model:  gemma3-4b 1.08 (n=300)   gpt-oss-20b 1.35 (n=300)   gpt-oss-120b 1.51 (n=299)
  finalize:        off 1.30 (n=449)   on 1.32 (n=450)

best configuration: finalize=on | generate.model=gpt-oss-120b | retrieve=dense_rerank  (mean 1.78)

============================================================
INFERENTIAL -- mixed-effects model (significance & effect sizes)
============================================================
verdict ~ (1 | input_id) + (retrieve + generate.model + finalize)^2
  fixed-effects model + Type-II ANOVA, up to 2-way   (n=899, alpha=0.05)

per-term effects  (F-test, p, partial eta^2;  'x' = interaction):
  term                             F          p    partial eta^2
  retrieve                    130.37     0.0000        0.228   ***
  generate.model               24.06     0.0000        0.052   ***
  retrieve x generate.model     4.17     0.0024        0.019   **
  generate.model x finalize     0.12     0.8858        0.000
  finalize                      0.17     0.6841        0.000
  retrieve x finalize           0.04     0.9612        0.000
  Signif. codes:  0 '***' 0.001 '**' 0.01 '*' 0.05 '.' 0.1 ' ' 1

effect sizes -- Cohen's d (magnitude of the gap; 0.2 small, 0.5 medium, 0.8 large):
  retrieve: dense vs dense_rerank              d = -0.31   95% CI [-0.47, -0.15]
  retrieve: dense vs none                      d = +0.86   95% CI [+0.69, +1.03]
  retrieve: dense_rerank vs none               d = +1.24   95% CI [+1.07, +1.42]
  generate.model: gemma3-4b vs gpt-oss-120b    d = -0.49   95% CI [-0.65, -0.33]
  generate.model: gemma3-4b vs gpt-oss-20b     d = -0.30   95% CI [-0.46, -0.14]
  generate.model: gpt-oss-120b vs gpt-oss-20b  d = +0.18   95% CI [+0.02, +0.34]
  finalize: off vs on                          d = -0.02   95% CI [-0.16, +0.11]

note: random intercept not estimable; using fixed-effects ANOVA
note: model was near-singular (little between-question variance) -- treat p-values as
      unstable; the effect sizes (Cohen's d) are the more reliable signal here

============================================================
ORDINAL -- cumulative-link mixed model (verdicts as ordered categories)
============================================================
ordinal CLMM -- verdict ~ (1 | input_id) + (retrieve + generate.model + finalize)^2
  (n=899, logLik=-557.7, alpha=0.05)

fixed effects (ordinal log-odds of a higher score; + = better):
  term                                                  estimate       p
  retrieve=dense_rerank                                   +1.149    0.0075   **
  retrieve=none                                           -3.944    0.0000   ***
  generate.model=gpt-oss-120b                             +1.250    0.0035   **
  generate.model=gpt-oss-20b                              +0.800    0.0541   .
  finalize=on                                             -0.070    0.8524
  retrieve=dense_rerank x generate.model=gpt-oss-120b     -0.146    0.7947
  retrieve=none x generate.model=gpt-oss-120b             +1.854    0.0006   ***
  retrieve=dense_rerank x generate.model=gpt-oss-20b      -0.041    0.9390
  retrieve=none x generate.model=gpt-oss-20b              +0.880    0.0936   .
  retrieve=dense_rerank x finalize=on                     +0.033    0.9418
  retrieve=none x finalize=on                             +0.089    0.8313
  generate.model=gpt-oss-120b x finalize=on               +0.160    0.7166
  generate.model=gpt-oss-20b x finalize=on                +0.302    0.4852
  Signif. codes:  0 '***' 0.001 '**' 0.01 '*' 0.05 '.' 0.1 ' ' 1
\end{lstlisting}

\end{document}